\documentclass[10pt,twocolumn,letterpaper]{article}

\usepackage{iccv}
\usepackage{times}
\usepackage{epsfig}
\usepackage{graphicx}
\usepackage{amsmath}
\usepackage{amssymb}
\usepackage{afterpage}
\usepackage{rotating}
\usepackage{xcolor,colortbl}
\usepackage{placeins}
\usepackage[outercaption]{sidecap}  
\usepackage{makecell}

\usepackage[pagebackref=true,breaklinks=true,letterpaper=true,colorlinks,bookmarks=false]{hyperref}

\iccvfinalcopy 

\def\iccvPaperID{1777} 

\begin{document}

\title{Unsupervised Visual Representation Learning by 
Context Prediction}

\author{
\vspace{-0.2in}
\begin{tabular}[t]{c @{\extracolsep{1em}} c @{\extracolsep{1em}} c}
        Carl Doersch$^{1,2}$ &
        Abhinav Gupta$^{1}$ &
        Alexei A. Efros$^{2}$ 
        \\
\end{tabular}
\cr
\cr
\small
\begin{tabular}[t]{c@{\extracolsep{4em}}c} 
       $^1$ School of Computer Science &
       $^2$ Dept. of Electrical Engineering and Computer Science \\
       Carnegie Mellon University &
       University of California, Berkeley \\
\end{tabular}
}
\maketitle

\begin{abstract}
This work explores the use of spatial context as a source of free and plentiful supervisory signal for training a rich visual representation.  Given only a large, unlabeled image collection, we extract random pairs of patches from each image and train a convolutional neural net to predict the position of the second patch relative to the first. We argue that doing well on this task requires the model to learn to recognize objects and their parts. We demonstrate that the feature representation learned
using this within-image context indeed captures visual similarity across images. For example, this representation allows us to perform unsupervised visual discovery of objects like cats, people, and even birds from the Pascal VOC 2011 detection dataset. Furthermore, we show that the learned ConvNet can be used in the R-CNN framework~\cite{girshick2014rich} and provides a significant boost over a randomly-initialized ConvNet, resulting in state-of-the-art performance among algorithms which use only Pascal-provided training set annotations.

\end{abstract}

\vspace{-0.2in}
\section{Introduction}
\vspace{-0.05in}
Recently, 
new computer vision methods have leveraged large datasets of millions of labeled examples to learn rich, high-performance visual representations~\cite{krizhevsky2012imagenet}. 
Yet efforts to scale these methods to truly Internet-scale datasets (i.e. hundreds of {\bf b}illions of images) are hampered by the sheer expense of the human annotation required. 
A natural way to address this difficulty would be to employ unsupervised learning, which aims to use data without any annotation.  Unfortunately, despite several decades of sustained effort, unsupervised methods
have not yet been shown to extract useful information from
large collections of full-sized, real images.  
After all, without labels, it is not even clear \textit{what} should be represented. How can one write an objective function to encourage a representation to capture, for example, objects, if none of the objects are labeled?   

\begin{figure}[t]
\begin{center}

   \includegraphics[width=0.9\linewidth]{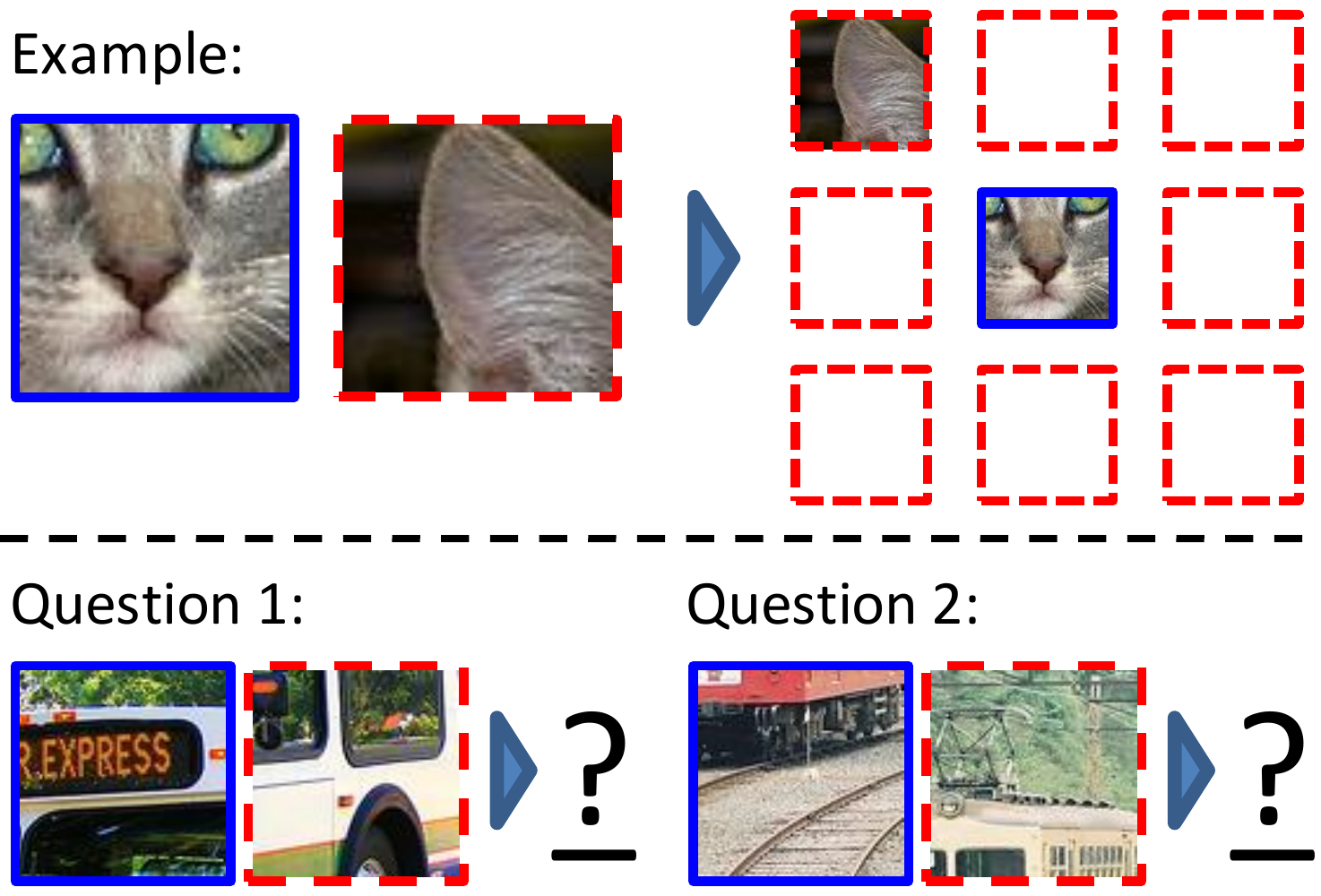}
   \vspace{-.2cm}
\end{center}
   \caption{Our task for learning patch representations involves randomly sampling a patch (blue) and then one of eight possible neighbors (red).  Can you guess the spatial configuration for the two pairs of patches?  Note that the task is much easier once you have recognized the object! 
}
\hfill\protect\rotatebox[origin=c]{180}{\begin{small}Answer key: Q1: Bottom right Q2: Top center\end{small}}
\vspace{-.2in}
\label{fig:quiz}
\end{figure}


Interestingly, in the text domain, \textit{context} has proven to be a powerful source of automatic supervisory signal for learning representations~\cite{ando2005framework,tsujiiythu2007discriminative,collobert2008unified,mikolov2013distributed}. 
Given a large text corpus, the idea is to train a model that maps each word to a feature vector, such that it is easy to predict the words in the context (i.e., a few words before and/or after) given the vector.  This converts an apparently unsupervised problem (finding a good similarity metric between words) into a 
``self-supervised'' one: learning a function from a given word to the words surrounding it. Here the context prediction task is just a ``pretext'' to force the model to learn a good word embedding, which, in turn, has been shown to be useful in a number of real tasks, such as semantic word similarity~\cite{mikolov2013distributed}.

Our paper aims to provide a similar ``self-supervised'' formulation for image data: a supervised task involving predicting the context for a patch. Our task 
is illustrated in Figures~\ref{fig:quiz} and~\ref{fig:task}.  We sample random pairs of patches in one of eight spatial configurations, and present each pair to a machine learner, providing no information about the patches' original position within the image.  The algorithm must then guess the position of one patch relative to the other.  Our underlying hypothesis is that doing well on this task requires understanding scenes and objects, {\em i.e.} a good visual representation for this task will need to extract objects and their parts in order to reason about their relative spatial location. ``Objects,'' after all, consist of multiple parts that can be detected independently of one another, and which occur in a specific spatial configuration (if there is no specific configuration of the parts, then it is  ``stuff''~\cite{adelson2001seeing}).
We present a ConvNet-based approach to learn a visual representation from this task. We demonstrate that the resulting visual representation is good for both object detection, providing a significant boost on PASCAL VOC 2007 compared to learning from scratch, as well as for unsupervised object discovery / visual data mining.  This means, surprisingly, that our representation generalizes {\em across} images, despite being trained using an objective function that operates on a single image at a time.   That is, instance-level supervision appears to improve performance on category-level tasks.


\vspace{-0.05in}
\section{Related Work}
\vspace{-0.05in}
One way to think of a good image representation is as the latent variables of an appropriate generative model. 
An ideal generative model of natural images would both generate images according to their natural distribution, and be concise in the sense that it would seek common causes for different images and share information between them.
However, inferring the latent structure given an image is intractable for even relatively simple models.
To deal with these computational issues, a number of works, such as the wake-sleep algorithm~\cite{hinton1995wake}, contrastive divergence~\cite{hinton2006fast}, deep Boltzmann machines~\cite{salakhutdinov2009deep}, and variational Bayesian methods~\cite{kingma2014,rezende2014stochastic} use sampling to perform approximate inference. Generative models have shown promising performance on smaller datasets such as handwritten digits~\cite{hinton1995wake,hinton2006fast,salakhutdinov2009deep,kingma2014,rezende2014stochastic}, but none have proven effective for high-resolution natural images.




Unsupervised representation learning can also be formulated as learning an embedding (i.e. a feature vector for each image) where images that are semantically similar are close, while semantically different ones are far apart. 
One way to build such a representation is to create a supervised ``pretext'' task such that an embedding which solves the task will also be useful for other real-world tasks.  For example, denoising autoencoders~\cite{vincent2008extracting,bengio2013deep} use reconstruction from noisy data as a pretext task: the algorithm must connect images to other images with similar objects to tell the difference between noise and signal. Sparse autoencoders also use reconstruction as a pretext task, along with a sparsity penalty~\cite{olshausen1996emergence}, and such autoencoders may be stacked to form a deep representation~\cite{lee2006efficient,le2013building}. 
(however, only~\cite{le2013building} was successfully applied to full-sized images, requiring a million CPU hours to discover just three objects). 
We believe that current reconstruction-based algorithms struggle with low-level phenomena, like stochastic textures, making it hard to even measure whether a model is generating well.  




Another pretext task is ``context prediction.''
A strong tradition for this kind of task already exists in the text domain, where ``skip-gram''~\cite{mikolov2013distributed} models have been shown to generate useful word representations.  The idea is to train a model (e.g. a deep network) to predict, from a single word, the $n$ preceding and $n$ succeeding words.  In principle, similar reasoning could be applied in the image domain, a kind of visual ``fill in the blank'' task, but, again, one runs into the problem of determining whether the predictions themselves are correct~\cite{doersch2014context},
unless one cares about predicting only very low-level features~\cite{domke2008killed,larochelle2011neural,theis2015generative}.
To address this,~\cite{malisiewicz2009beyond} predicts the appearance of an image region by consensus voting of the transitive nearest neighbors of its surrounding regions. Our previous work~\cite{doersch2014context} explicitly formulates a statistical test to determine whether 
the data is better explained by a prediction or by a low-level null hypothesis model. 


\begin{figure}[t]
\begin{center}
   \includegraphics[width=0.9\linewidth]{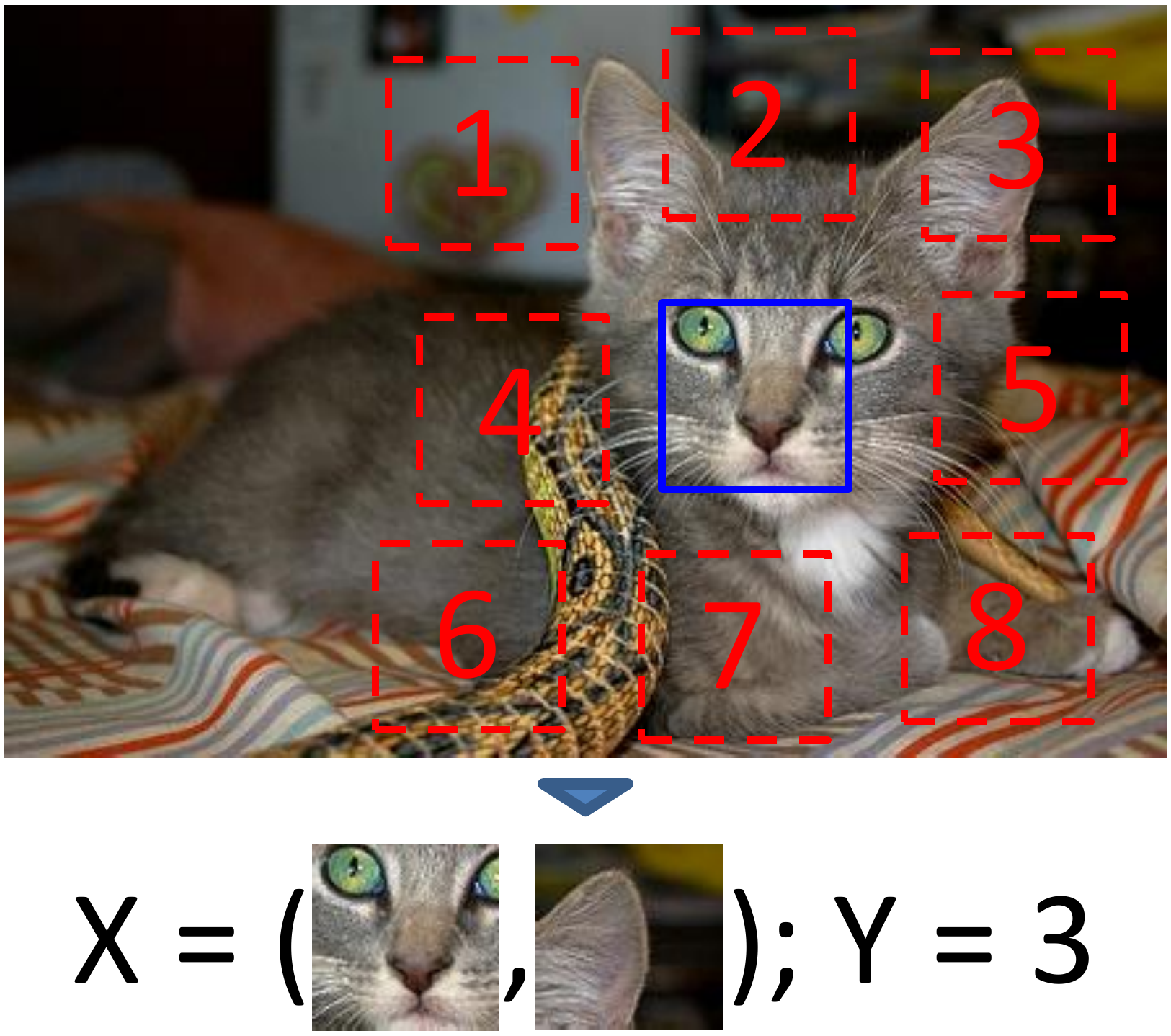}
   \vspace{-.2cm}
\end{center}
   \caption{The algorithm receives two patches in one of these eight possible spatial arrangements, without any context, and must then classify which configuration was sampled. }
   \vspace{-.2in}
\label{fig:task}
\end{figure}


The key problem that these approaches must address is that predicting pixels is much harder than predicting words, due to the huge variety of pixels that can arise from the same semantic object. In the text domain, one 
interesting idea is to switch from a pure prediction task to a discrimination task~\cite{tsujiiythu2007discriminative,collobert2008unified}. 
In this case, the pretext task is to discriminate true snippets of text from the same snippets where a word has been replaced at random.  
A direct extension of this to 2D might be to discriminate between real images vs. images where one patch has been replaced by a random patch from elsewhere in the dataset. 
However, such a task would be trivial, since discriminating low-level color statistics and lighting would be enough.  To make the task harder and more high-level,
in this paper, we instead classify between multiple possible configurations of patches sampled from {\em the same image}, which means they will share lighting and color statistics, as shown on Figure~\ref{fig:task}.

Another line of work in unsupervised learning from images aims to discover object categories using hand-crafted features and various forms of clustering (e.g.~\cite{sivic2005discovering,russell2006using} learned a generative model over bags of visual words). 
Such representations lose shape information, and will readily discover clusters of, say, foliage.   A few subsequent works have attempted to use representations more closely tied to shape \cite{lee2009foreground,payet2010set}, but relied on contour extraction, which is difficult in complex images.  Many other approaches~\cite{grauman2006unsupervised,kim2008unsupervised,faktor2012clustering} focus on defining similarity metrics which can be used in more standard clustering algorithms; ~\cite{RematasCVPR15}, for instance, re-casts the problem as frequent itemset mining. Geometry may also be used to for verifying links between images~\cite{quack2008world,Chum09,heath2010image}, although this can fail for deformable objects.

Video can provide another cue for representation learning.  For most scenes, the identity of objects remains unchanged even as appearance changes with time.  This kind of temporal coherence has a long history in visual learning literature~\cite{foldiak1991learning,wiskott02}, and contemporaneous work shows strong improvements on modern detection datasets~\cite{wang2015unsupervised}.

Finally, our work is related to a line of research on discriminative patch
mining~\cite{doersch2012makes,singh2012unsupervised,juneja13blocks,li2013harvesting,sun2013learning,doersch2013mid}, which has emphasized weak supervision as a means of object discovery.  Like the current work, they emphasize the utility of learning representations of patches (i.e. object parts) before learning full objects and scenes, and argue that scene-level labels can serve as a pretext task.  
For example, \cite{doersch2012makes} trains detectors to be sensitive to different geographic locales, but the actual goal 
is to discover specific elements of architectural style.

\vspace{-0.05in}
\section{Learning Visual Context Prediction}\label{sec:learning}
\vspace{-0.05in}
We aim to learn an image representation for our pretext task, i.e., predicting the relative position of patches within an image.  We employ Convolutional Neural Networks (ConvNets), which are well known to learn complex image representations with minimal human feature design.
Building a ConvNet that can predict a relative offset for a pair of patches is, in principle, straightforward: the network must feed the two input patches through several convolution layers, and produce an output that assigns a probability to each of the eight spatial configurations (Figure~\ref{fig:task}) that might have been sampled (i.e. a softmax output).  Note, however, that we ultimately wish to learn a feature embedding for {\em individual} patches, such that patches which are visually similar (across different images) would be close in the 
embedding space.  

\begin{figure}[t]
\begin{center}

   \includegraphics[width=0.8\linewidth]{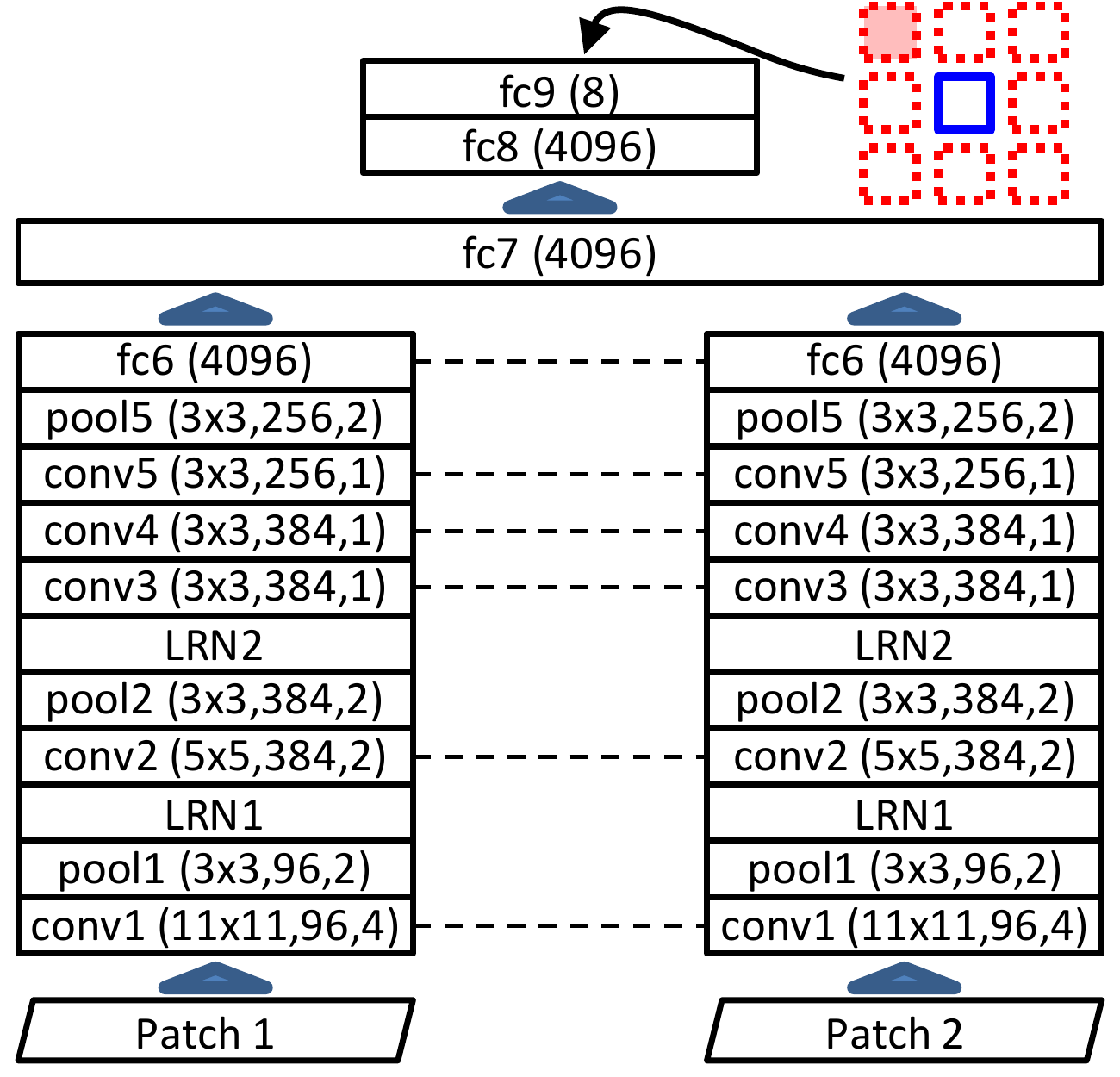}
   \vspace{-.1in}
\end{center}
   \caption{Our architecture for pair classification.  Dotted lines indicate shared weights.  `conv' stands for a convolution layer, `fc' stands for a fully-connected one, `pool' is a max-pooling layer, and `LRN' is a local response normalization layer.  Numbers in parentheses are kernel size, number of outputs, and stride (fc layers have only a number of outputs).  The LRN parameters follow~\cite{krizhevsky2012imagenet}.  All conv and fc layers are followed by ReLU nonlinearities, except fc9 which feeds into a softmax classifier. }
   \vspace{-.2in}
\label{fig:arch}
\end{figure}

To achieve this, we use a late-fusion architecture shown in Figure~\ref{fig:arch}: a pair of AlexNet-style architectures~\cite{krizhevsky2012imagenet} that process each patch separately, until a depth analogous to fc6 in AlexNet, after which point the representations are fused.  For the layers that process only one of the patches, weights are tied between both sides of the network, such that the same fc6-level embedding function is computed for both patches.  
Because there is limited capacity for joint reasoning---i.e., only two layers receive input from both patches---we expect the network to perform the bulk of the semantic reasoning for each patch separately.
When designing the network, we followed AlexNet where possible.  

To obtain training examples given an image, we sample the first patch uniformly, without any reference to image content.
Given the position of the first patch, we sample the second patch randomly from the eight possible neighboring locations as in Figure~\ref{fig:task}.  

\vspace{-0.05in}
\subsection{Avoiding ``trivial'' solutions}
\vspace{-0.05in}

When designing a pretext task, care must be taken to ensure that the task forces the network to extract the desired information (high-level semantics, in our case), 
without taking ``trivial'' shortcuts. 
In our case, low-level cues like boundary patterns or textures continuing between patches could potentially serve as such a shortcut.  
Hence, for the relative prediction task, it was important to include a gap between patches (in our case, approximately half the patch width).  
Even with the gap, it is possible that long lines spanning neighboring patches could could give away the correct answer.
Therefore, we also randomly jitter each patch location by up to 7 pixels (see Figure~\ref{fig:task}).






\begin{figure*}[t]
\begin{center}

   \includegraphics[width=0.85\linewidth]{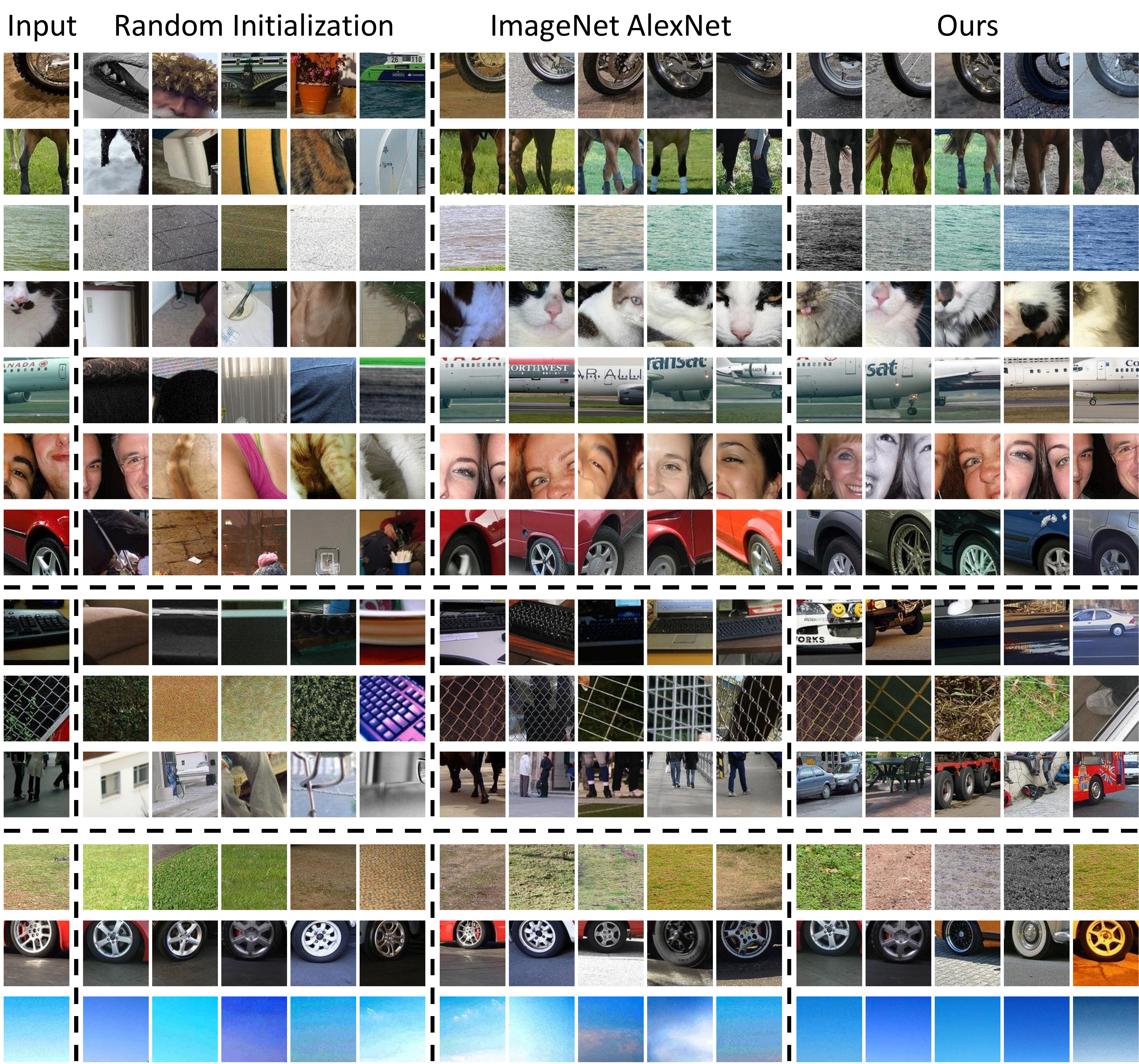}
   \vspace{-.1in}
\end{center}
   \caption{Examples of patch clusters obtained by nearest neighbors.
   The query patch is shown on the far left.  Matches are for three different features: fc6 features from a random initialization of our architecture, AlexNet fc7 after training on labeled ImageNet, and the fc6 features learned from our method.  Queries were chosen from 1000 randomly-sampled patches.  The top group is examples where our algorithm performs well; for the middle AlexNet outperforms our approach; and for the bottom all three features work well.}
   \vspace{-.2in}
\label{fig:nns}
\end{figure*}

However, even these precautions are not enough:
we were surprised to find that, for some images, another trivial solution exists.  We traced the problem to an unexpected culprit: chromatic aberration.
Chromatic aberration arises from differences in the way the lens focuses light at different wavelengths.
In some cameras, one color channel (commonly green) is shrunk toward the image center relative to the others~\cite[p.~76]{brewster1854treatise}.
A ConvNet, it turns out, can learn to localize a patch relative to the lens itself (see Section~\ref{sec:aberration}) simply by detecting the separation between green and magenta (red + blue).  Once the network learns the absolute location on the lens, solving the relative location task becomes trivial.
To deal with this problem, we experimented with two types of pre-processing.  One is to shift green and magenta toward gray (`projection').  Specifically, let $a=[-1,2,-1]$ (the 'green-magenta color axis' in RGB space).  We then define $B=I-a^{T}a/(aa^{T})$, which is a matrix that subtracts the projection of a color onto the green-magenta color axis.  We multiply every pixel value by $B$.
An alternative approach is to randomly drop 2 of the 3 color channels from each patch (`color dropping'), replacing the dropped colors with Gaussian noise (standard deviation $\sim 1/100$ the standard deviation of the remaining channel). For qualitative results, we show the `color-dropping' approach, but found both performed similarly; for the object detection results, we show both results.

\begin{figure*}[t]
\begin{center}
   \includegraphics[width=0.99\linewidth]{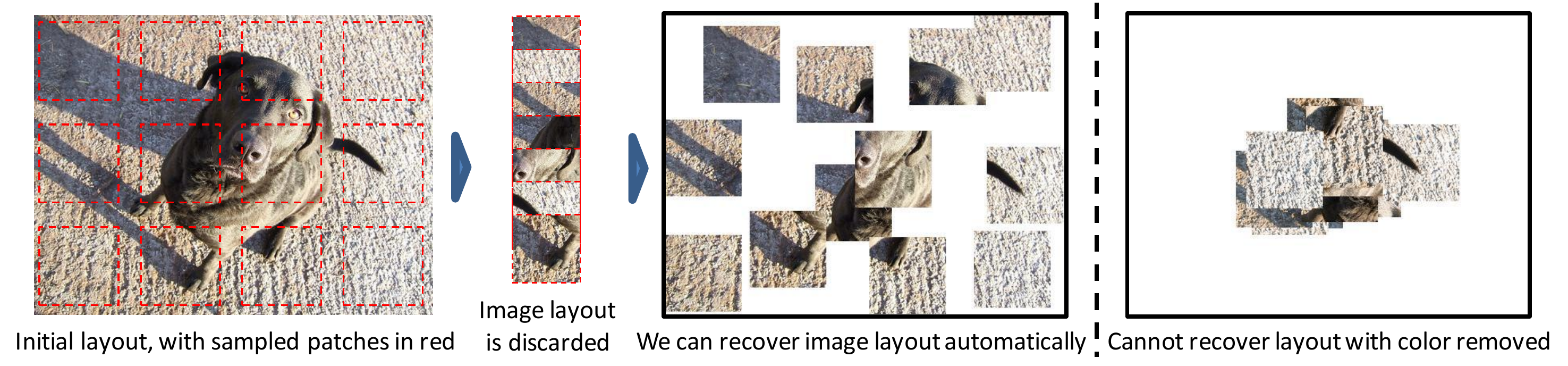}
\end{center}
    \vspace{-0.2in}
   \caption{
    We trained a network to predict the absolute $(x,y)$ coordinates of randomly sampled patches.
    Far left: input image. 
    Center left: extracted patches.
    Center right: the location the trained network predicts for each patch shown on the left.  Far right: the same result after our color projection scheme.  Note that the far right patches are shown \textit{after} color projection; the operation's effect is almost unnoticeable.
    }
    \vspace{-0.2in}
\label{fig:aberration}
\end{figure*}

\noindent {\bf Implementation Details:} We use Caffe~\cite{jia2014caffe}, and train on the ImageNet~\cite{deng2009imagenet} 2012 training set (~1.3M images), using only the images and discarding the labels. First, we resize each image to between 150K and 450K total pixels, preserving the aspect-ratio. From these images, we sample patches at resolution 96-by-96. For computational efficiency, we only sample the patches from a grid like pattern, such that each sampled patch can participate in as many as 8 separate pairings. We allow a gap of 48 pixels between the sampled patches in the grid, but also jitter the location of each patch in the grid by $-7$ to $7$ pixels in each direction.
We preprocess patches by (1) mean subtraction
(2) projecting or dropping colors (see above), and (3) randomly downsampling some patches to as little as 100 total pixels, and then upsampling it, to build robustness to pixelation. When applying simple SGD to train the network, we found that the network predictions would degenerate to a uniform prediction over the 8 categories, with all activations for fc6 and fc7 collapsing to 0. This meant that the optimization became permanently stuck in a saddle point where it ignored the input from the lower layers (which helped minimize the variance of the final output), and therefore that the net could not tune the lower-level features and escape the saddle point.  Hence, our final implementation employs batch normalization~\cite{ioffe2015batch}, without the scale and shift ($\gamma$ and $\beta$), which forces the network activations to vary across examples.  We also find that high momentum values (e.g. $.999$) accelerated learning.  For experiments, we use a ConvNet trained on a K40 GPU for approximately four weeks.  


\vspace{-0.1in}
\section{Experiments}
\vspace{-0.05in}

We first demonstrate the network has learned to associate semantically similar patches, using simple nearest-neighbor matching.  We then apply the trained network in two domains.  
First, we use the model as ``pre-training'' for a standard vision task with only limited training data: specifically, we use the VOC 2007 object detection.  Second, we evaluate visual data mining, where the goal is to start with an unlabeled image collection and discover object classes. 
Finally, we analyze the performance on the layout prediction ``pretext task'' to see how much is left to learn from this supervisory signal.

\vspace{-0.05in}
\subsection{Nearest Neighbors}\label{sec:nns}
\vspace{-0.05in}
Recall our intuition that training should assign similar representations to semantically similar patches.  In this section, our goal is to understand which patches our network considers similar.
We begin by sampling random 96x96 patches, which we represent using fc6 features (i.e. we remove fc7 and higher shown in Figure~\ref{fig:arch}, and use only one of the two stacks).  We find nearest neighbors using normalized correlation of these features.  Results for some patches (selected out of 1000 random queries) are shown in Figure~\ref{fig:nns}.
For comparison, we repeated the experiment using fc7 features from AlexNet trained on ImageNet (obtained by upsampling the patches), and using fc6 features from our architecture but without any training (random weights initialization).  As shown in Figure~\ref{fig:nns}, the matches returned by our feature often capture the semantic information that we are after, matching AlexNet in terms of semantic content (in some cases, e.g. the car wheel, our matches capture pose better). Interestingly, in a few cases, random (untrained) ConvNet also does reasonably well.

\vspace{-0.05in}
\subsection{Aside: Learnability of Chromatic Aberration}\label{sec:aberration}
\vspace{-0.05in}
We noticed in early nearest-neighbor experiments that some patches retrieved match patches from the same absolute location in the image, regardless of content, because those patches displayed similar aberration.  
To further demonstrate this phenomenon, we trained a network 
to predict the absolute $(x,y)$ coordinates of patches sampled from ImageNet. While the overall accuracy of this regressor is not very high, it does surprisingly well for some images: for the top 10\% of images, the average (root-mean-square) error is .255, while chance performance (always predicting the image center) yields a RMSE of .371.  Figure~\ref{fig:aberration} shows one such result. 
Applying the proposed ``projection'' scheme increases the error on the top 10\% of images to .321.

\begin{table*}
\scriptsize{
\setlength{\tabcolsep}{3pt}
\center
\definecolor{LightRed}{rgb}{1,.5,.5}
\begin{tabular}{c|c c c c c c c c c c c c c c c c c c c c|c}
\hline
\hline
\textbf{VOC-2007 Test}&
      aero &        bike &        bird &        boat &        bottle &        bus &        car &        cat &        chair &        cow &        table &        dog &        horse &        mbike &        person &        plant &        sheep &        sofa &        train &        tv &        mAP\\
\hline
\hline
\textbf{DPM-v5}\cite{felzenszwalb2010}&        
        33.2 &        60.3 &        10.2 &        16.1 &        27.3 &        54.3 &        58.2 &        23.0 &        20.0 &        24.1 &        26.7 &        12.7 &        58.1 &        48.2 &        43.2 &        12.0 &        21.1 &        36.1 &        46.0 &        43.5 &        33.7\\
\hline
\textbf{\cite{cinbis2013segmentation}} w/o context &        
       52.6 &        52.6 &        19.2 &        25.4 &        18.7 &        47.3 &        56.9 &        42.1 &        16.6 &        41.4 &        41.9 &        27.7 &        47.9 &        51.5 &        29.9 &        20.0 &        41.1 &        36.4 &        48.6 &        53.2 &        38.5\\
\hline
\textbf{Regionlets\cite{WangRegionlets}}&        
       54.2 &        52.0 &        20.3 &        24.0 &        20.1 &         55.5 &        68.7 &        42.6 &        19.2 &        44.2 &        49.1 &        26.6 &        57.0 &        54.5 &        43.4 &        16.4 &        36.6 &       37.7 &        59.4 &        52.3 &        41.7 \\
\hline
\textbf{Scratch-R-CNN\cite{agrawal2014analyzing}}&        
        49.9 &        60.6 &        24.7 &        23.7 &        20.3 &        52.5 &        64.8 &        32.9 &        20.4 &        43.5 &        34.2 &        29.9 &        49.0 &       60.4 &        47.5 &        28.0 &        42.3 &        28.6 &        51.2 &        50.0 &        40.7\\
\hline
\textbf{Scratch-Ours}&
       52.6 &        60.5 &        23.8 &        24.3 &        18.1 &        50.6 &        65.9 &        29.2 &        19.5 &        43.5 &        35.2 &        27.6 &        46.5 &        59.4 &        46.5 &        25.6 &        42.4 &        23.5 &        50.0 &        50.6 &        39.8\\
\hline
\textbf{Ours-projection}&        
        58.4 &        62.8 &        33.5 &        27.7 &        24.4 &        58.5 &        68.5 &        41.2 &        26.3 &        49.5 &        42.6 &        37.3 &        55.7 &        62.5 &        49.4 &        29.0 &        47.5 &        28.4 &        54.7 &         56.8 &        45.7\\
\hline
\textbf{Ours-color-dropping}&
        60.5 &        66.5 &        29.6 &        28.5 &        26.3 &        56.1 &        70.4 &        44.8 &        24.6 &        45.5 &        45.4 &        35.1 &        52.2 &        60.2 &         50.0 &        28.1 &        46.7 &        42.6 &        54.8 &        58.6 &        46.3\\
\hline
\textbf{Ours-Yahoo100m}&
        56.2 &        63.9 &        29.8 &        27.8 &        23.9 &        57.4 &        69.8 &        35.6 &        23.7 &        47.4 &        43.0 &        29.5 &        52.9 &        62.0 &        48.7 &        28.4 &        45.1 &        33.6 &        49.0 &        55.5 &        44.2\\
\hline
\hline
\textbf{ImageNet-R-CNN\cite{girshick2014rich}}&        
        64.2 &        69.7 &        50 &        41.9 &        32.0 &        62.6 &        71.0 &        60.7 &        32.7 &        58.5 &        46.5 &        56.1 &        60.6 &        66.8 &        54.2 &        31.5 &        52.8 &        48.9 &        57.9 &        64.7 &        54.2\\
\hline
\hline
\textbf{K-means-rescale~\cite{krahenbuhl2015data}}&
55.7 &        60.9 &        27.9 &        30.9 &        12.0 &        59.1 &        63.7 &        47.0 &        21.4 &        45.2 &        55.8 &        40.3 &        67.5 &        61.2 &        48.3 &        21.9 &        32.8 &        46.9 &        61.6 &        51.7 &        45.6 \\
\hline
\textbf{Ours-rescale~\cite{krahenbuhl2015data}}&
61.9 &        63.3 &        35.8 &        32.6 &        17.2 &        68.0 &        67.9 &        54.8 &        29.6 &        52.4 &        62.9 &        51.3 &        67.1 &        64.3 &        50.5 &        24.4 &        43.7 &        54.9 &        67.1 &        52.7 &        51.1\\
\hline
\textbf{ImageNet-rescale~\cite{krahenbuhl2015data}}&
64.0 &        69.6 &        53.2 &        44.4 &        24.9 &        65.7 &        69.6 &        69.2 &        28.9 &        63.6 &        62.8 &        63.9 &        73.3 &        64.6 &        55.8 &        25.7 &        50.5 &        55.4 &        69.3 &        56.4 &        56.5 \\
\hline
\hline
\textbf{VGG-K-means-rescale}&
56.1 &        58.6 &        23.3 &        25.7 &        12.8 &        57.8 &        61.2 &        45.2 &        21.4 &        47.1 &        39.5 &        35.6 &        60.1 &        61.4 &        44.9 &        17.3 &        37.7 &        33.2 &        57.9 &        51.2 &        42.4 \\
\hline
\textbf{VGG-Ours-rescale}&
71.1 &        72.4 &        54.1 &        48.2 &        29.9 &        75.2 &        78.0 &        71.9 &        38.3 &        60.5 &        62.3 &        68.1 &        74.3 &        74.2 &        64.8 &        32.6 &        56.5 &        66.4 &        74.0 &        60.3 &        61.7\\
\hline
\textbf{VGG-ImageNet-rescale}&
76.6 &        79.6 &        68.5 &        57.4 &        40.8 &        79.9 &        78.4 &        85.4 &        41.7 &        77.0 &        69.3 &        80.1 &        78.6 &        74.6 &        70.1 &        37.5 &        66.0 &        67.5 &        77.4 &        64.9 &        68.6 \\
\hline
\hline

\end{tabular}
\caption{Mean Average Precision on VOC-2007.}
\label{tab:voc_2007}
\vspace{-.25cm}
}
\end{table*}

\vspace{-0.05in}
\subsection{Object Detection}
\vspace{-0.05in}
\label{sec:obj_det}

Previous work
on the Pascal VOC challenge~\cite{everingham2010pascal}
has shown that pre-training on ImageNet (i.e., training a ConvNet to solve the ImageNet challenge) and then ``fine-tuning'' the network (i.e. re-training the ImageNet model for PASCAL data) provides a substantial boost over training on the Pascal training set alone~\cite{girshick2014rich,agrawal2014analyzing}.  However, as far as we are aware, no works have shown that \textit{unsupervised} pre-training on images can provide such a performance boost, no matter how much data is used.

\begin{figure}
  \begin{minipage}[c]{0.4\linewidth}
    \includegraphics[width=\textwidth]{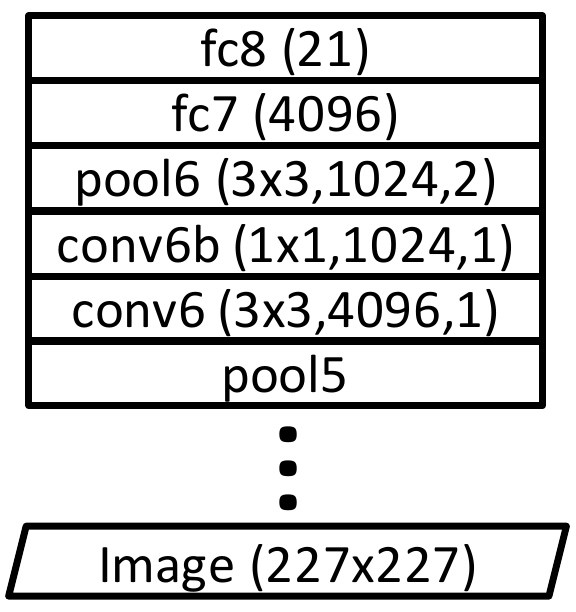}
  \end{minipage}\hfill
  \begin{minipage}[c]{0.57\linewidth}
    \caption{Our architecture for Pascal VOC detection.  Layers from conv1 through pool5 are copied from our patch-based network (Figure~\ref{fig:arch}).  The new 'conv6' layer is created by converting the fc6 layer into a convolution layer.  Kernel sizes, output units, and stride are given in parentheses, as in Figure~\ref{fig:arch}.} \label{fig:fcarch}
  \end{minipage}
  \vspace{-.6cm}
\end{figure}

Since we are already using a ConvNet, we adopt the current state-of-the-art R-CNN pipeline~\cite{girshick2014rich}.  R-CNN works on object proposals that have been resized to 227x227.  Our algorithm, however, is aimed at 96x96 patches.  We find that downsampling the proposals to 96x96 loses too much detail.
Instead, we adopt the architecture shown in Figure~\ref{fig:fcarch}.  As above, we use only one stack from Figure~\ref{fig:arch}.
Second, we resize the convolution layers to operate on inputs of 227x227.  This results in a pool5 that is 7x7 spatially, so we must convert the previous fc6 layer into a convolution layer  (which we call conv6) following~\cite{long2014fully}.  Note our conv6 layer has 4096 channels, where each unit connects to a 3x3 region of pool5.  A conv layer with 4096 channels would be quite expensive to connect directly to a 4096-dimensional fully-connected layer.  Hence, we add another layer after conv6 (called conv6b), using a 1x1 kernel, which reduces the dimensionality to 1024 channels (and adds a nonlinearity).   Finally, we feed the outputs through a pooling layer to a fully connected layer (fc7) which in turn connects to a final fc8 layer which feeds into the softmax.  We fine-tune this network according to the procedure described in~\cite{girshick2014rich} (conv6b, fc7, and fc8 start with random weights), and use fc7 as the final representation.  We do not use bounding-box regression, and take the appropriate results from~\cite{girshick2014rich} and~\cite{agrawal2014analyzing}.

Table~\ref{tab:voc_2007} shows our results.  Our architecture trained from scratch (random initialization) performs slightly worse than AlexNet trained from scratch.  
However, our pre-training makes up for this, boosting the from-scratch number by 6\% MAP, and outperforms an AlexNet-style model trained from scratch on Pascal by over 5\%.  This puts us about 8\% behind
the performance of R-CNN pre-trained with ImageNet labels~\cite{girshick2014rich}.  This is the best result we are aware of on VOC 2007 without using labels outside the dataset.  We ran additional baselines initialized with batch normalization, but found they performed worse than the ones shown.

To understand the effect of various dataset biases~\cite{torralba11}, we also performed a preliminary experiment pre-training on a randomly-selected 2M subset of the Yahoo/Flickr 100-million Dataset~\cite{thomee2015yfcc100m}, which was collected entirely automatically.  The performance after fine-tuning is slightly worse than Imagenet, but there is still a considerable boost over the from-scratch model.

In the above fine-tuning experiments, we removed the batch normalization layers by estimating the
mean and variance of the conv- and fc- layers, and then rescaling the weights and biases such that the outputs of the conv and fc layers have mean 0 and variance 1 for each channel.  
Recent work~\cite{krahenbuhl2015data},
however, has shown empirically that the scaling of the weights prior to finetuning can have a
strong impact on test-time performance, and argues that our previous method of
removing batch normalization leads too poorly scaled weights.  They propose a simple way to
rescale the network's weights without changing the function that the network computes, such that
the network behaves better during finetuning.  Results using this technique are shown 
in Table~\ref{tab:voc_2007}.
Their approach gives a boost to all methods, but gives less of a boost to the 
already-well-scaled ImageNet-category model.  Note that for this comparison, we 
used fast-rcnn~\cite{girshickICCV15fastrcnn} to save compute time, and we discarded all 
pre-trained fc-layers from our model, re-initializing them with the K-means procedure 
of~\cite{krahenbuhl2015data}
(which was used to initialize all layers in the ``K-means-rescale'' row).  
Hence, the structure of the network during
fine-tuning and testing was the
same for all models.

Considering that we have essentially infinite data to train our model, we might expect
that our algorithm should also provide a large boost to higher-capacity models such as 
VGG~\cite{Simonyan14c}.  To test this, we trained a model following the 16-layer
structure of~\cite{Simonyan14c} for the convolutional layers on each side of the network 
(the final fc6-fc9 layers were the same as in Figure~\ref{fig:arch}).  
We again
fine-tuned the representation on Pascal VOC using fast-rcnn, by transferring only the 
conv layers, again following Kr{\"a}henb{\"u}hl et al.~\cite{krahenbuhl2015data} to 
re-scale the transferred weights and
initialize the rest.  As a baseline, we performed a similar experiment with the ImageNet-pretrained
16-layer model of~\cite{Simonyan14c} (though we kept pre-trained fc layers
rather than re-initializing them), 
and also by initializing the entire network with
K-means~\cite{krahenbuhl2015data}.  Training time was considerably longer---about 8 weeks
on a Titan X GPU---but the the network outperformed the AlexNet-style model by a considerable
margin.  Note the model initialized with K-means performed roughly on par with the analogous
AlexNet model, suggesting that most of the boost came from the unsupervised pre-training.


\begin{figure*}
\begin{center}
   \includegraphics[width=0.95\linewidth]{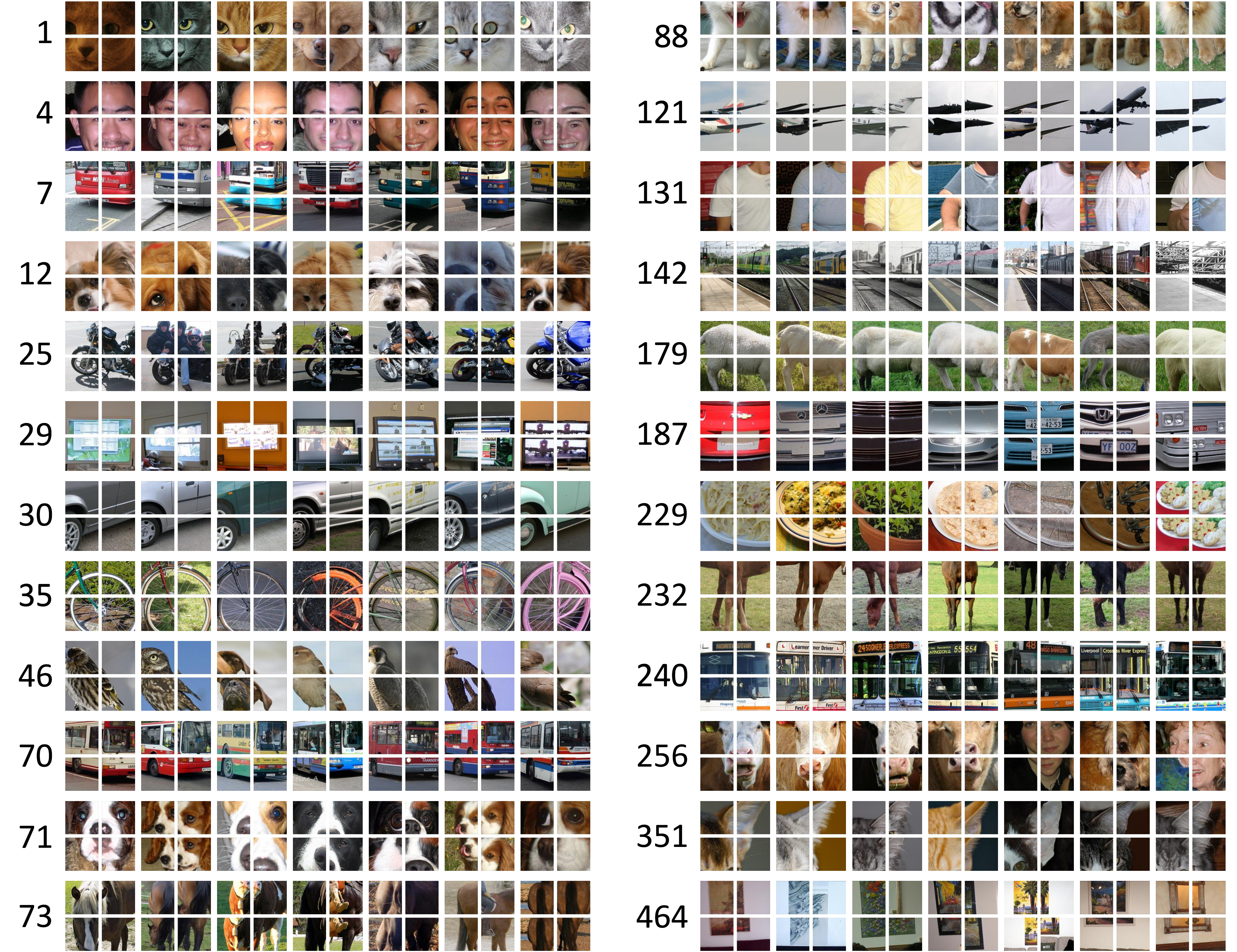}
\end{center}
\vspace{-0.1in}
   \caption{Object clusters discovered by our algorithm.  The number beside each cluster indicates its ranking, determined by the fraction of the top matches that geometrically verified.  
   For all clusters, we show the raw top 7 matches that verified geometrically.  
   The full ranking is available on our project webpage.}
   \vspace{-0.2in}
\label{fig:discovered}
\end{figure*}

\begin{table}

\setlength{\tabcolsep}{3pt}
\center
\definecolor{LightRed}{rgb}{1,.5,.5}
\begin{tabular}{l c c c c c}
\Xhline{2\arrayrulewidth}
 &\multicolumn{2}{c}{Lower Better} & \multicolumn{3}{c}{Higher Better}\\
& Mean & Median & $11.25^{\circ}$ & $22.5^{\circ}$ & $30^{\circ}$\\
\hline
Scratch & 38.6 & 26.5 & 33.1 & 46.8 & 52.5 \\
Unsup. Tracking~\cite{wang2015unsupervised} & 34.2 & 21.9 & 35.7 & 50.6 & 57.0 \\
Ours & \textbf{33.2} & 21.3 & 36.0 & 51.2 & 57.8 \\
ImageNet Labels & 33.3 & \textbf{20.8} & \textbf{36.7} & \textbf{51.7} & \textbf{58.1} \\
\Xhline{2\arrayrulewidth}
\end{tabular}
\vspace{.05cm}
\caption{Accuracy on NYUv2.}
\label{tab:surf_norm}
\vspace{-.25cm}

\end{table}

\vspace{-0.05in}
\subsection{Geometry Estimation}
\vspace{-0.05in}

The results of Section~\ref{sec:obj_det} suggest that our representation is sensitive 
to objects, even though it was not originally trained to find them.  This raises the question:
Does our representation extract information that is useful for other, non-object-based tasks?
To find out, we fine-tuned our network to perform the surface normal estimation on NYUv2 proposed in Fouhey et al.~\cite{Fouhey13a}, following the finetuning procedure of Wang et al.~\cite{wang2015unsupervised} (hence, we compare directly to the unsupervised pretraining results
reported there).  We used the color-dropping network, restructuring the fully-connected
layers as in Section~\ref{sec:obj_det}.  Surprisingly, our results are almost equivalent to
those obtained using a fully-labeled ImageNet model.
One possible explanation for this is that the ImageNet categorization task does relatively
little to encourage a network to pay attention to geometry, since the geometry is largely
irrelevant once an object is identified.  Further evidence of this can be seen in seventh row of
Figure~\ref{fig:nns}: the nearest neighbors for ImageNet AlexNet are all car wheels, but they are
not aligned well with the query patch.

\vspace{-0.05in}
\subsection{Visual Data Mining}
\label{sec:datamining}
\vspace{-0.05in}
Visual data mining~\cite{quack2008world,doersch2012makes,singh2012unsupervised,RematasCVPR15}, or unsupervised object discovery~\cite{sivic2005discovering,russell2006using,grauman2006unsupervised}, 
aims to use a large image collection to discover image fragments which happen to depict the same semantic objects. 
Applications include dataset visualization, content-based retrieval, and tasks that require relating visual data to other unstructured information (e.g. GPS coordinates~\cite{doersch2012makes}).
For automatic data mining, our approach from section~\ref{sec:nns} is inadequate: 
although object patches match to similar objects, textures match just as readily to similar textures.  Suppose, however, that we sampled two non-overlapping patches from the same object.  Not only would the nearest neighbor lists for both patches share many images, but within those images, the nearest neighbors would be in roughly the same spatial configuration.  For texture regions, on the other hand, the spatial configurations of the neighbors would be random, because the texture has no global layout.

To implement this, we first sample a constellation of four adjacent patches from an image (we use four to reduce the likelihood of a matching spatial arrangement happening by chance). 
We find the top 100 images which have the strongest matches for all four patches, ignoring spatial layout. 
We then use a type of geometric verification~\cite{chum2007total} to filter away the images where the four matches are not geometrically consistent.
Because our features are more semantically-tuned, we can use a much weaker type of geometric verification than~\cite{chum2007total}.  
Finally, we rank the different constellations by counting the number of times the top 100 matches geometrically verify.

\noindent {\bf Implementation Details:} To compute whether a set of four matched patches geometrically verifies, we first compute the best-fitting square $S$ to the patch centers (via least-squares), while constraining that side of $S$ be between $2/3$ and $4/3$ of the average side of the patches.  We then compute the squared error of the patch centers relative to $S$ (normalized by dividing the sum-of-squared-errors by the square of the side of $S$).  The patch is geometrically verified if this normalized squared error is less than $1$.  When sampling patches do not use any of the data augmentation preprocessing steps (e.g. downsampling).  We use the color-dropping version of our network. 

We applied the described mining algorithm to Pascal VOC 2011, with no pre-filtering of images and no additional labels.
We show some of the resulting patch clusters in Figure~\ref{fig:discovered}.  The results are visually comparable to our previous work~\cite{doersch2014context}, although we discover a few objects that were not found in~\cite{doersch2014context}, such as monitors, birds, torsos, and plates of food.  The discovery of birds and torsos---which are notoriously deformable---provides further evidence for the invariances our algorithm has learned.  
We believe we have covered all objects discovered in~\cite{doersch2014context}, with the exception of (1) trusses and (2) railroad tracks without trains (though we do discover them with trains).  For some objects like dogs, we discover more variety and rank the best ones higher.  Furthermore, many of the clusters shown in~\cite{doersch2014context} depict gratings (14 out of the top 100), whereas none of ours do (though two of our top hundred depict diffuse gradients).  As in~\cite{doersch2014context}, we often re-discover the same object multiple times with different viewpoints, which accounts for most of the gaps between ranks in Figure~\ref{fig:discovered}.  The main disadvantages of our algorithm relative to~\cite{doersch2014context} are 1) some loss of purity, and 2) that we cannot currently determine an object mask automatically (although one could imagine dynamically adding more sub-patches to each proposed object).

\begin{figure}[t]
\begin{center}

   \includegraphics[width=0.9\linewidth]{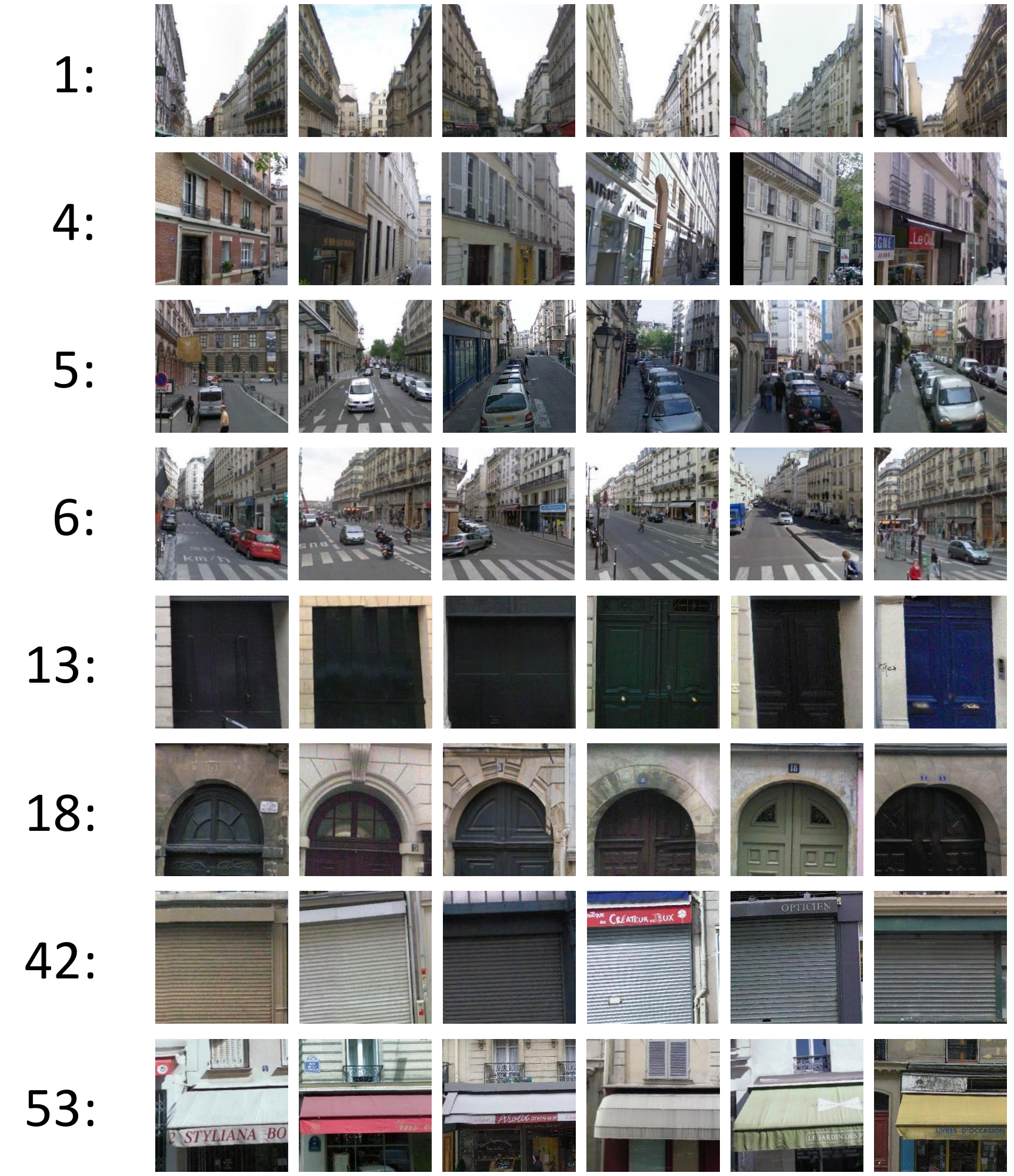}
   \vspace{-.2cm}
\end{center}
\vspace{-0.05in}
   \caption{Clusters discovered and automatically ranked via our algorithm (\S~\ref{sec:datamining}) from the Paris Street View dataset. }
   \vspace{-0.1in}
\label{fig:paris}
\end{figure}

To ensure that our algorithm has not simply learned an object-centric representation due to the various biases~\cite{torralba11} in ImageNet, we also applied our algorithm to 15,000 Street View images from Paris (following~\cite{doersch2012makes}).  The results in Figure~\ref{fig:paris} show that our representation captures scene layout and architectural elements. For this experiment, to rank clusters, we use the de-duplication procedure originally proposed in~\cite{doersch2012makes}.

\vspace{-0.15in}
\subsubsection{Quantitative Results} \label{quantitative}
\begin{figure}[t]
\begin{center}
   \includegraphics[width=0.8\linewidth]{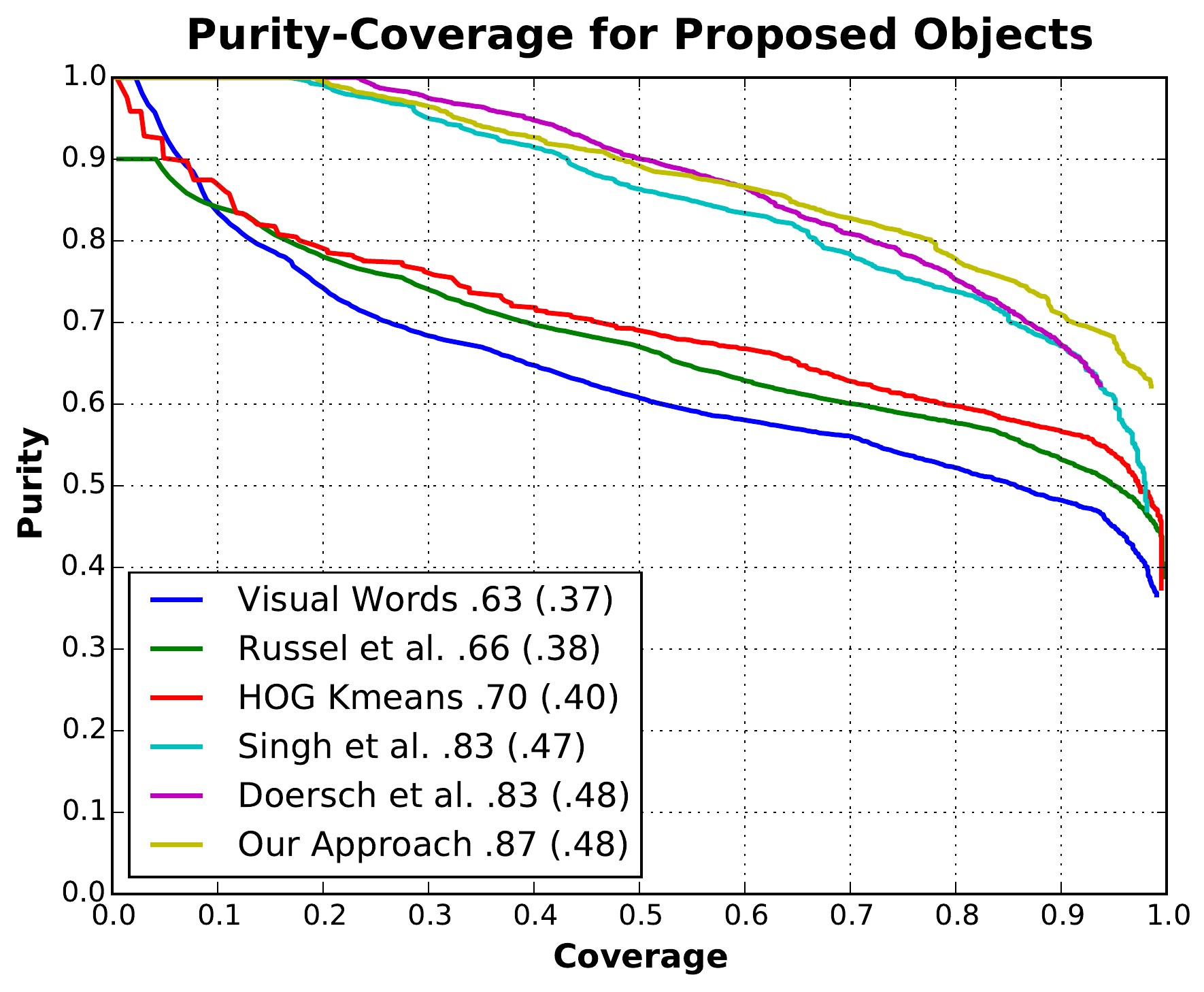}
   \vspace{-.2in}
\end{center}
   \caption{Purity vs coverage for objects discovered on a subset of Pascal VOC 2007. The numbers in the legend indicate area under the curve (AUC). In parentheses is the AUC up to a coverage of .5.}
   \vspace{-.2in}
\label{fig:purcov}
\end{figure}

\vspace{-0.05in} 
As part of the qualitative evaluation, we applied our algorithm to the subset of Pascal VOC 2007 selected in~\cite{singh2012unsupervised}: specifically, those containing at least one instance of \textit{bus}, \textit{dining table}, \textit{motorbike}, \textit{horse}, \textit{sofa}, or \textit{train}, and evaluate via a purity coverage curve following~\cite{doersch2014context}.  We select 1000 sets of 10 images each for evaluation.  The evaluation then sorts the sets by \textit{purity}: the fraction of images in the cluster containing the same category.  We generate the curve by walking down the ranking.  For each point on the curve, we plot average purity of all sets up to a given point in the ranking against \textit{coverage}: the fraction of images in the dataset that are contained in at least one of the sets up to that point.  As shown in Figure~\ref{fig:purcov}, we have gained substantially in terms of coverage, suggesting increased invariance for our learned feature.  However, we have also lost some highly-pure clusters compared to~\cite{doersch2014context}---which is not very surprising considering that our validation procedure is considerably simpler.

\noindent {\bf Implementation Details:} We initialize 16,384 clusters by sampling patches, mining nearest neighbors, and geometric verification ranking as described above.  The resulting clusters are highly redundant.
The cluster selection procedure of~\cite{doersch2014context} relies on a likelihood ratio score that is calibrated across clusters, which is not available to us.  
To select clusters, we first select the top 10 geometrically-verified neighbors for each cluster.  Then we iteratively select the highest-ranked cluster that contributes at least one image to our coverage score.  When we run out of images that aren't included in the coverage score, we choose clusters to cover each image at least twice, and then three times, and so on.

\subsection{Accuracy on the Relative Prediction Task Task}\label{pretext}
Can we improve the representation by further training on our relative prediction pretext task?  To find out, we briefly analyze classification performance on pretext task itself.
We sampled 500 random images from Pascal VOC 2007, sampled 256 pairs of patches from each, and classified them into the eight relative-position categories from Figure~\ref{fig:task}.  This gave an accuracy of 38.4\%, where chance performance is 12.5\%, suggesting that the pretext task is quite hard (indeed, human performance on the task is similar).  
To measure possible overfitting, we also ran the same experiment on ImageNet, which is the dataset we used for training.  The network was 39.5\% accurate on the training set, and 40.3\% accurate on the validation set (which the network never saw during training), suggesting that little overfitting has occurred.

One possible reason why the pretext task is so difficult is because, for a large fraction of patches within each image, the task is almost impossible.  Might the task be easiest for image regions corresponding to objects? To test this hypothesis, we repeated our experiment using only patches sampled from within Pascal object ground-truth bounding boxes.  We select only those boxes that are at least 240 pixels on each side, and which are not labeled as truncated, occluded, or difficult.  Surprisingly, this gave essentially the same accuracy of 39.2\%, 
and a similar experiment only on cars yielded 45.6\% accuracy.   So, while our algorithm is sensitive to objects, it is almost as sensitive to the layout of the rest of the image.

\footnotesize \noindent {\bf Acknowledgements} We thank Xiaolong Wang and Pulkit Agrawal for help with baselines,
Berkeley and CMU vision group members for many fruitful discussions, and Jitendra Malik for putting gelato on the line. This work was partially supported by Google Graduate Fellowship to CD, ONR MURI N000141010934, Intel research grant, an NVidia hardware grant, and an Amazon Web Services grant.




{\footnotesize
\bibliographystyle{ieee}
\bibliography{egbib}
}

\end{document}